\documentclass[letterpaper]{article} % DO NOT CHANGE THIS
\usepackage{aaai25}
\usepackage{times}  % DO NOT CHANGE THIS
\usepackage{helvet}  % DO NOT CHANGE THIS
\usepackage{courier}  % DO NOT CHANGE THIS
\usepackage[hyphens]{url}  % DO NOT CHANGE THIS
\usepackage{graphicx} % DO NOT CHANGE THIS
\usepackage{natbib}  % DO NOT CHANGE THIS AND DO NOT ADD ANY OPTIONS TO IT
\usepackage{caption} % DO NOT CHANGE THIS AND DO NOT ADD ANY OPTIONS TO IT
\usepackage{algorithm}
\usepackage{algorithmic}

\usepackage{newfloat}
\usepackage{listings}
\DeclareCaptionStyle{ruled}{labelfont=normalfont,labelsep=colon,strut=off} % DO NOT CHANGE THIS
\floatstyle{ruled}
\newfloat{listing}{tb}{lst}{}
\floatname{listing}{Listing}
\usepackage{amssymb}
\usepackage{multirow}

\title{Graffin: Stand for Tails in Imbalanced Node Classification}
\author{
    Xiaorui Qi,
    Yanlong Wen\textsuperscript{*}, 
    Xiaojie Yuan
}
\affiliations{
    College of Computer Science, Nankai University \\
    qixiaorui@mail.nankai.edu.com, \{wenyl, yuanxj\}@nankai.edu.cn
}

\begin{document}

\maketitle

\begin{abstract}
Graph representation learning (GRL) models have succeeded in many scenarios.
Real-world graphs have imbalanced distribution, such as node labels and degrees, which leaves a critical challenge to GRL.
Imbalanced inputs can lead to imbalanced outputs.
However, most existing works ignore it and assume that the distribution of input graphs is balanced, which cannot align with real situations, resulting in worse model performance on tail data.
The domination of head data makes tail data underrepresented when training graph neural networks (GNNs).
Thus, we propose \textbf{Graffin}, a pluggable tail data augmentation module, to address the above issues.
Inspired by recurrent neural networks (RNNs), Graffin flows head features into tail data through graph serialization techniques to alleviate the imbalance of tail representation.
The local and global structures are fused to form the node representation under the combined effect of neighborhood and sequence information, which enriches the semantics of tail data.
We validate the performance of Graffin on four real-world datasets in node classification tasks.
Results show that Graffin can improve the adaptation to tail data without significantly degrading the overall model performance.
\end{abstract}


\section{Introduction}
Graph representation learning (GRL) models have succeeded in many scenarios, like recommendations~\cite{ref:rec}, bioinformatics~\cite{ref:SMILES}, etc.
Node classification is one of the typical tasks~\cite{ref:nc0,ref:nc1,ref:nc2}, where the goal is to learn a classifier predicting the ground labels of nodes.
Researchers have explored graph neural networks (GNNs) in this domain, proposing amounts of foundation models~\cite{ref:GCN,ref:GAT,ref:GraphSMOTE}.
Existing works achieve noteworthy performance, most assuming that the distribution is balanced.

\begin{figure}[!t]
    \centering
    \includegraphics[width=\linewidth]{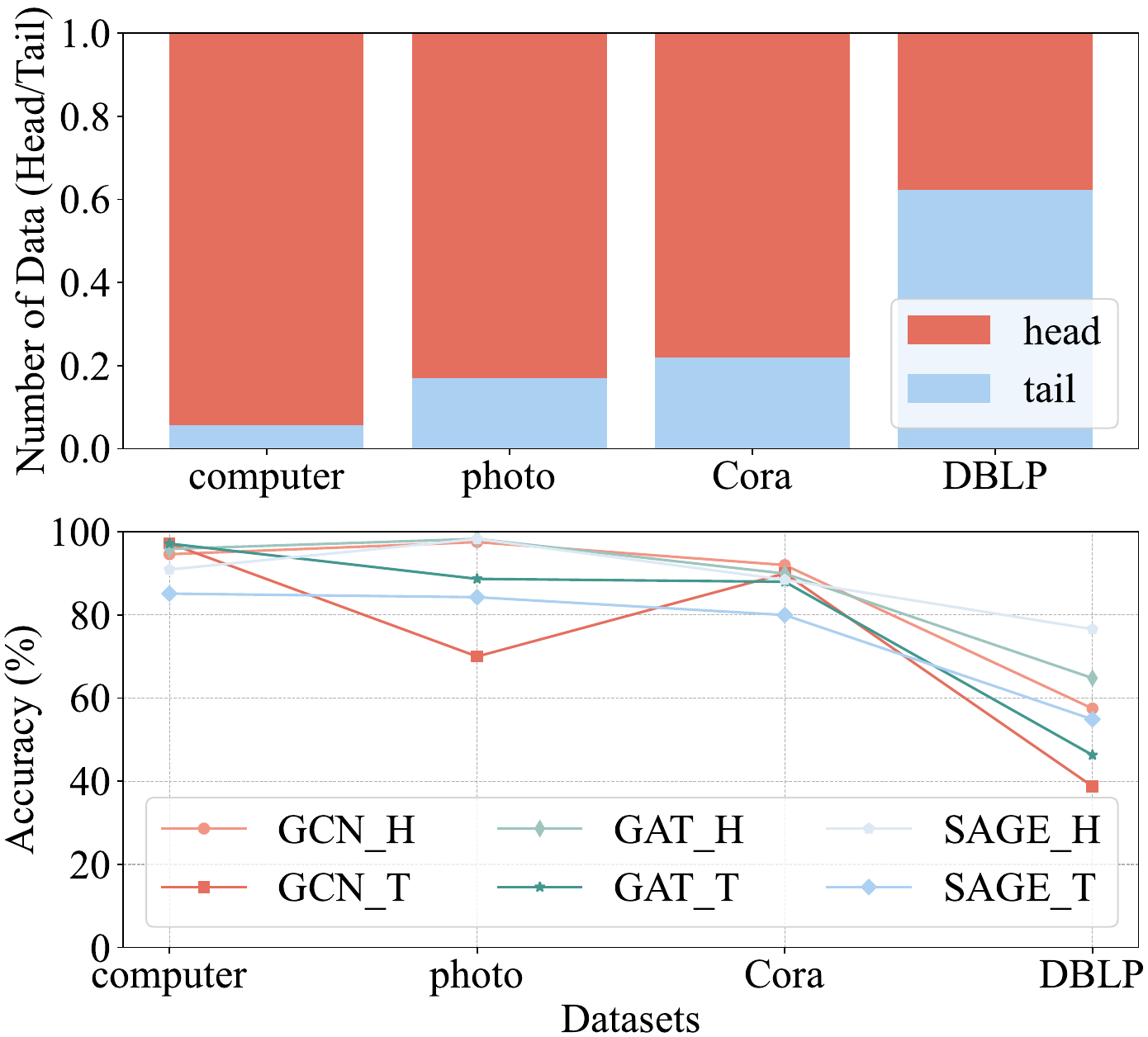}
    \caption{Imbalanced characteristics in node classification tasks. \textbf{Top}: Imbalanced distribution of node classes. \textbf{Bottom}: Imbalanced classification performance, where H and T denote the results of head and tail data.}
    \label{fig:pre}
\end{figure}
However, real-world graphs usually have imbalanced distribution, such as node labels and degrees, which leaves a critical challenge to GRL.
For example, in Figure~\ref{fig:pre}, we illustrate the imbalanced statistics of four real-world datasets.
Each bar shows the proportion of the tail data (blue) to the head data (red).
The tails are far less than the heads.
In the worst case, the tail data is only 5.6\% of the head.
Though DBLP has a relatively well-balanced situation, with fewer classes having less gap between each other, tails are still 38\% less than head data.
Imbalanced inputs can lead to imbalanced outputs.
We further train three vanilla models to test the classification performance, and the results show that models perform better on the head.
We take \emph{photo} as an example, with 1941 head data and 331 tail data.
The average model performances are $98.06 \pm 0.36$ and $81.01 \pm 7.97$, respectively.
Head data are better represented and more robust, having a superiority of about 17.05\%.
The domination of head data makes tail data underrepresented and turbulent, resulting in worse tail classification accuracy.

This phenomenon brings out an interesting topic of \emph{enhancing the performance of GRL models, especially for tail data, in an imbalanced setting}.
Researchers have worked on imbalanced graph learning (IGL)~\cite{ref:GraphSMOTE,ref:ENS} for a long time.
They improve the tail data performance or balance heads and tails through sampling or generating strategies.
However, these approaches change the original graph distribution and may introduce noise due to virtual nodes and edges.
We focus on building a structure-oriented strategy with only original graph structural information to alleviate the noise caused by sampling.
Some works~\cite{ref:LTE4G,ref:G2GNN} have tried augmentation approaches, using specific knowledge or information to enrich the tails.
It suffers from obtaining complicated structures like subgraphs with unbearable cost.

Moreover, existing works consider the imbalance under a single scenario like node class or degree.
Nevertheless, sources causing imbalance are various and intertwined in real-world cases~\cite{ref:Imb}.
For example, ordinary users on social networks account for the vast majority (a large sample number in a particular label), but each of them owns poor interests (low degree).
On the contrary, a small portion of core users can absorb several times the number of connections other users have.
It urges the proposal of a mixed framework that combines various imbalance factors.

Therefore, in this paper, we proposed a novel pluggable module, \textbf{Graffin}, to stand for tails and improve their performance in imbalanced node classification.
Our goal is to enhance the adaptation to tail data without significantly degrading the overall model performance.
We introduce an information augmentation strategy instead of sampling that changes the original distribution.
We flow more context to tail data via the graph serialization technique, orientating to the graph structure without generating new nodes and edges.
Next, we design a mixup of local and global structures, which enriches the semantics of tail data to the same level as the heads.
It is a combination framework of GNN and RNN, where the former aggregates 1-hop neighbors through the message-passing mechanism to represent local features, and the latter learns sequential information from heads to tails as a sample of global features.
Most importantly, the information flow of the graph sequence is in serialization order, jumping out of the limitation of topological structures.
It could learn latent features from long-distance node pairs.
Diversity in both features balances the fusion, forming the final node representation.

In summary, our main contributions are as follows:
\begin{itemize}
    \item As far as we know, we are the first to integrate sequential information to graph data under an imbalanced scenario. We flow head features into tail data through graph serialization techniques instead of sampling to alleviate the imbalance of tail representation. It orientates toward the original graphs without changing the distribution or generating synthetic nodes.
    \item We propose a novel pluggable module, \textbf{Graffin}, leveraging sequential semantics via RNNs. We fuse the local and global structures to form the node representation under the combined effect of neighborhood and sequence information, which enriches the semantics of tail data.
    \item We validate the performance of Graffin on four real-world datasets in node classification tasks. Results show that Graffin can improve the adaptation to tail data without significantly degrading the overall model performance. Moreover, other classes, not only tails, can benefit from the sequence.
\end{itemize}

\section{Related Work}
\subsection{Imbalanced Graph Learning}
Imbalanced graph learning (IGL)~\cite{ref:ImGCL,ref:QTIAN,ref:Imb} derives from traditional imbalanced learning works and is more complicated since graphs have various sources of imbalance, such as classes, degrees, etc.
The data-level algorithm is one of the mainstream countermeasures for solving imbalanced problems.
Inspired by SMOTE~\cite{ref:SMOTE}, GraphSMOTE~\cite{ref:GraphSMOTE} synthesizes nodes for tail classes, oversampling until balanced.
It constructs new edges between existing nodes and synthetic samples.
GraphENS~\cite{ref:ENS} takes a further step, generating the neighbors of tail data rather than only themselves.
It builds subgraphs of tails to provide more semantics.
CM-GCL~\cite{ref:CMGCL} puts imbalanced learning into a contrastive learning paradigm, developing a co-modality framework to generate contrastive pairs automatically.
However, virtual nodes and edges sampled and generated in the works above change the original graph distribution, leading to extra noise.

Despite the re-balancing approaches of the GraphSMOTE branch, other works~\cite{ref:GraphMixup,ref:LTE4G,ref:G2GNN} aim to enrich tail data with extra information augmentation.
GraphMixup~\cite{ref:GraphMixup}, inspired by Mixup, designs multiple augmentation strategies from various aspects.
It can adaptively change the upsampling ratio of tail data with the power of reinforcement learning.
G2GNN~\cite{ref:G2GNN} adds a graph-of-graph view based on kernel similarity to obtain high-level supernode features on graph tasks.
Meanwhile, ImGCL~\cite{ref:ImGCL} also tries contrastive learning, leveraging the progressively balanced sampling method with pseudo-labels.
These works use extra information augmentation, keeping alignment with the original graph distribution.
But they suffer to compute more complex structures such as subgraphs, graph-of-graphs, etc.

For more details about IGL, we suggest reading the surveys~\cite{ref:survey2023,ref:survey2024}.
\subsection{Graph Sequence Modeling}
GNNs can handle neighbors well but fail to capture long-distance node interactions due to their message-passing mechanism.
However, models such as recurrent neural networks (RNNs)~\cite{ref:GRU} are powerful in dealing with sequential data.
They can learn rich representation and remember historical information in a long sequence.
The non-serializable property is one of the vital resistances of applying these models to graphs.
Though some works attempt to combine graphs and sequences, most of which are rich in sequential semantics, such as time series in temporal graphs~\cite{ref:time} and SMILES sequence in molecular graphs~\cite{ref:SMILES}.
Few methods are ready for general graphs to capture latent features behind sequences.
Recently, researchers have aimed at linear-time sequence modeling methods.
Many designs arise, such as selective state spaces~\cite{ref:SSMS}, real-gated linear recurrent units~\cite{ref:Griffin}, etc.
It allows rethinking the combination of graphs and sequences.

\begin{figure*}[!ht]
    \centering
    \includegraphics[width=\linewidth]{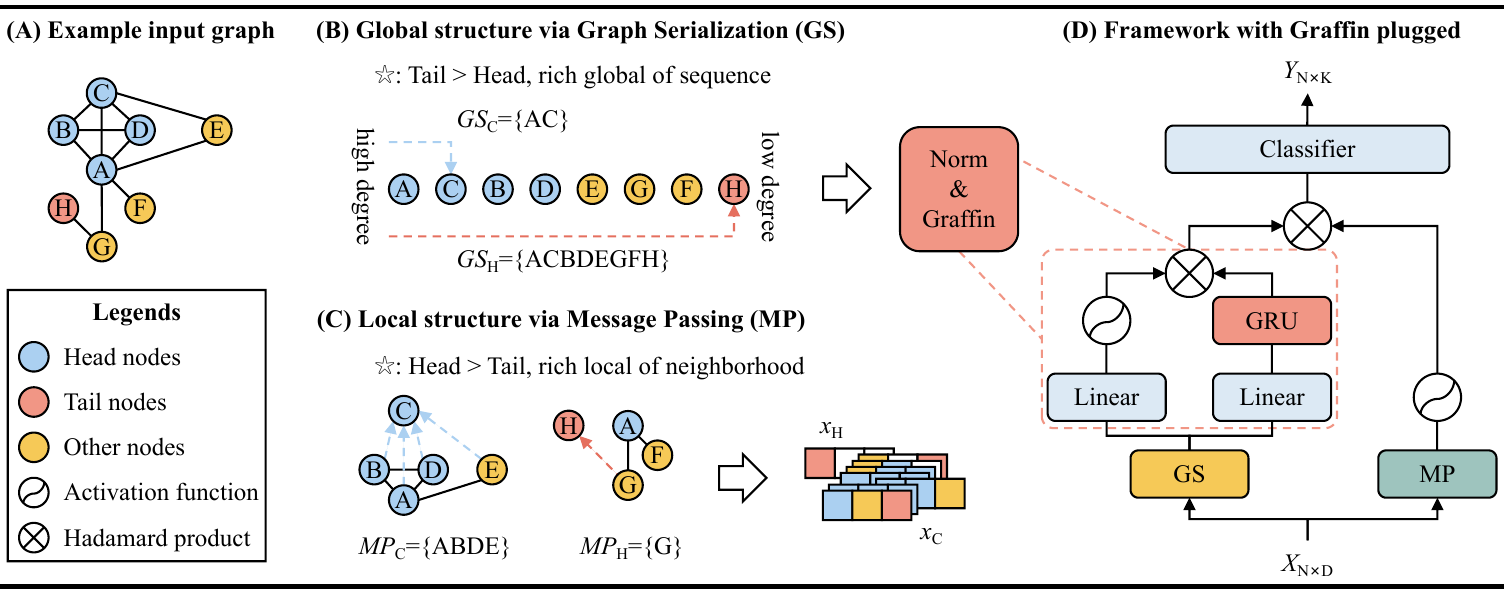}
    \caption{Overview of the proposed Graffin method. (A) Example input graph with imbalanced node class distribution. (B) Sequential global structure via graph serialization (GS). (C) 1-hop local structure via message passing (MP) neural networks. (D) Example framework with exchangeable MP with Graffin plugged.}
    \label{fig:framework}
\end{figure*}
\section{Preliminaries}
\paragraph{Attributed Graph.}
An attributed graph can form as $\mathcal{G} = \{\mathcal{V}, \mathcal{E}, \mathcal{X}, \Phi, \mathcal{Y}\}$, where $\mathcal{V} = \{v_1, ..., v_N\}$ and $\mathcal{E} = \{e_{ij} | v_i, v_j \in \mathcal{V}\}$ represent the set of nodes and edges, and $\mathcal{X} \in \mathbb{R}^{N \times D}$ denotes the graph attributes with $D$ dimensions.
The function $\Phi: \mathcal{V} \rightarrow \mathcal{Y}$ maps a node $v \in \mathcal{V}$ to its corresponding label $y \in \mathcal{Y}$.
We use $\mathcal{A} \in \{0, 1\}^{N \times N}$ to represent the adjacency matrix, where $\mathcal{A}[i, j] = 1$ if $e_{ij} \in \mathcal{E}$.
\paragraph{Imbalance Ratio.}
Imbalanced data distribution occurs frequently due to the complex characteristics of graphs.
In this paper, we aim at node class imbalance.
We evaluate it via a node class imbalance ratio, $R_{imb}$.
Mathematically, for a graph $\mathcal{G}$ with $K$ independent node classes, where $\mathcal{Y} = \bigcup_{1 \leq i \leq K} \mathcal{Y}_i$, the imbalance ratio is defined as
\begin{equation}
    R_{imb} = \frac{|\mathcal{Y}_{\max}|}{|\mathcal{Y}_{\min}|},
\end{equation}
where $|\mathcal{Y}_{\min}|$ and $|\mathcal{Y}_{\max}|$ denote the number of tail and head data.
$R_{imb} \geq 1$, and the larger the number, the more imbalanced the datasets.
It is an ideal balance when $R_{imb} = 1$.
\paragraph{Problem Formulation.}
Node classification is one of the most vital applications in GNNs, aiming at mapping all nodes into several classes.
Given a graph $\mathcal{G} = \{\mathcal{V}, \mathcal{E}, \mathcal{X}, \Phi, \mathcal{Y}\}$, the goal is to learn a new mapping function $\bar{\Phi}: \mathcal{V} \rightarrow \bar{\mathcal{Y}}$, predicting ground truths $\mathcal{Y}$ correctly.

\section{Graffin: The Proposed Framework}
Figure~\ref{fig:framework} shows the overview framework of our proposed method, Graffin.
Inspired by RNNs, we flow head features into tail data through graph serialization techniques to alleviate the imbalance of tail representation.
Heads own more neighborhoods than tails.
Thus, tails need to grab more in sequence.
We fuse the local and global structures to form the node representation under the combined effect of neighborhood and sequence information, which enriches the semantics of tail data.
\subsection{Graph Serialization}
\paragraph{Global Structure via Graph Serialization.}
One of the most popular methods for alleviating the imbalance is re-balancing~\cite{ref:GraphSMOTE,ref:ENS}.
However, these approaches change the original distribution of nodes.
It may introduce additional noise and can't align with reality.
With this in mind, we hope to find an information augmentation strategy instead of breaking the original distribution.
Some intuitive works~\cite{ref:LTE4G,ref:G2GNN} have focused on it.
However, they keep the balancing phase without being oriented to the original graph and have far more complex extra information (e.g., subgraphs) to compute.

Inspired by the concept of meta-path~\cite{ref:meta-path}, we serialize graph data to get global structures hiding in the graph sequence.
Given a node set $\mathcal{V}$ and its attribute $\mathcal{X}$, we have the graph serialization (GS) block as
\begin{equation}
    (\mathcal{V}', \mathcal{X}') = \textit{GS}^\sigma(\mathcal{V}, \mathcal{X}),
    \label{eq:GS}
\end{equation}
where $\mathcal{V}'$ is the indices of rearranged nodes, and $\mathcal{X}'$ is the new stacked attributes.
$\sigma$ denotes the serialization strategy GS picks.
We default it to node degrees according to the intertwined characteristics of class and structure~\cite{ref:Imb}.

Countering traditional ideas that more important nodes should have more context~\cite{ref:Graph-Mamba}, we compute the sequence through GS to enrich the tails.
In other words, tail data should seize a longer node list.
In sequential data, each node only updates based on previous nodes from hidden states due to the nature of sequences.
For example, as shown in Figure~\ref{fig:framework}(B), we serialize the input graph in a descending degree order, where $\forall v_i, v_j \in \mathcal{V}$, if $i < j$, then $\sigma_i > \sigma_j$.
Let $\textit{GS}_v$ be the historical set node $v$ could see.
The last tail data $H$, having $N-1$ (self-omitted) previous nodes as context, is far more than the head $C$, which has only one node $A$ in the set.
In usual cases, we define a global structure enriching ratio,
\begin{equation}
    R_g = \frac{|\bar{\textit{GS}}_{head}|}{|\bar{\textit{GS}}_{tail}|},
\end{equation}
to express the improvement of the tail context.
Head data have a relatively small amount of context, approximating $|\mathcal{Y}_{\max}| / 2$, while tails are at the end of the sequence, having almost $N$ nodes selected in the hidden state.
To make a simplification, we set $R_g \approx R_{imb} / N$.
\paragraph{Graffin: Control Information Flows.}
The intuition behind Graffin is simple.
RNN's autoregressive ability can easily capture long-distance node interactions.
After acquiring the graph sequence, it learns global structures by serialization order, jumping out of the limitations of edges.
For the last tail data, we show the information flow following GRU~\cite{ref:GRU}:
\begin{equation}
    h_{tail} = g \odot h_{pre} + (1 - g) \odot s_{tail},
\end{equation}
where $g$ denotes a learnable recurrence gate, $pre$ and $s$ are the previous node of the tail, and the hidden state of the tail, respectively.

Note that \textit{pre} is the predecessor of the last tail data by serialization order.
It drops the topological restriction, finding long-distance node pairs in different ego networks and memorizing their representations.
Those historical nodes build the context of the last tail data, fused in the hidden state:
\begin{equation}
    s_{tail} = \textit{tanh}(h_{tail}W_{tail} + (s_{pre} \odot r)W_{pre}),
\end{equation}
where $r$ is another learnable gate function used for updating, $W$ is the corresponding weighted matrix, and \textit{tanh} is the hyperbolic tangent function.
Both gate functions follow the same format as
\begin{equation}
    f(h_{tail}) = \delta(h_{tail}W_{tail}^f + h_{pre}W_{pre}^f),
\end{equation}
where $\delta$ refers to the activation function such as sigmoid.

This behavior could learn two similar nodes not directly connected with edges.
For example, as shown in Figures~\ref{fig:framework}(A) and (B), $F$ rather than $G$ is the previous node of the tail data $H$ in the graph sequence.
That is because $F$ has a nearer relationship with $H$ defined by $\sigma$ (both have one node degree), where $|\sigma_F - \sigma_H| \leq |\sigma_G - \sigma_H|$.
In the same way, $A$ is the predecessor of the head data $C$, \textit{w.r.t.} $B$ and $D$, due to the overlap in the neighborhoods $\mathcal{N}_C = \{BDE\} \subseteq \mathcal{N}_A$ and outer linkages with node $E$, which is not head data.

Therefore, we design our Graffin module following Griffin~\cite{ref:Griffin}.
It takes the normalized graph sequence as the input and separates two branches.
The first branch is a plain linear layer with \textit{GeLU} as the activation, keeping the original node features:
\begin{equation}
    \mathcal{T}_1 = \textit{GeLU}(linear(\mathcal{X'})).
    \label{eq:Graffin1}
\end{equation}
The second branch applies an RNN-based core (e.g., GRU) to control the information flowing from heads to tails:
\begin{equation}
    \mathcal{T}_2 = \textit{GRU}(linear(\mathcal{X'})).
    \label{eq:Graffin2}
\end{equation}

Ultimately, we merge two branches to get the final representation of global sequential structures:
\begin{equation}
    \mathcal{H}_g = \mathcal{T}_1 \odot \mathcal{T}_2,
    \label{eq:Graffin3}
\end{equation}
which is a fusion of node representation $\mathcal{T}_1$ and sequential information $\mathcal{T}_2$.
The former ensures that self-features do not drop while the latter learns the graph sequence, avoiding over-smooth or indistinguishable.
\subsection{Structure Mixing}
\paragraph{Local Structure via Message Passing.}
Noteworthyly, GNNs achieve venerable impacts on node representations through their message passing (MP) techniques~\cite{ref:GCN}.
We utilize MP to aggregate 1-hop neighborhoods, forming local structures of each node.
In general, for a node $v$, a possible layer of MP is
\begin{equation}
    h_v = \textit{aggr}_{u \in \mathcal{N}(v)}(h_u),
    \label{eq:MP}
\end{equation}
where $\mathcal{N}(v)$ represents the neighbor nodes of $v$, and \textit{aggr} can be any aggregating functions (e.g., sum or average neighborhood features).
As shown in Figure~\ref{fig:framework}(C), for an imbalanced graph, head data own richer representations than tails due to various local structures of neighbors.
We use $\textit{MP}_v=\{u|u \in \mathcal{N}(v)\}$ to represent the neighbors of node $v$.
There are more nodes in the neighbors of the head data $C$ than the tail $H$, where $|\textit{MP}_C| = 4 \gg |\textit{MP}_A|$.
The more nodes the neighborhood has, the more representations the message-passing aggregates.
Therefore, it can define the imbalance in the local structure via the local ratio
\begin{equation}
    R_l = \frac{|\bar{\textit{MP}}_{head}|}{|\bar{\textit{MP}}_{tail}|}.
\end{equation}
Since tail data often have a relatively small number of neighbors, we can approximate $R_l$ with $N$ and $R_{imb}$, where $R_l \approx N / R_{imb}$.
\paragraph{Why Mixing Works.}
From the above discussion, we define two ratios, $R_g$ and $R_l$, where $R_gR_l \approx 1$.
It describes why combining local and global structures can enrich the representation of tail data.
$R_g$ flows the entire node set to tail data, showing the absolute position, while $R_l$ keeps the original neighborhood information as a relative one.
Thus, the fusion may balance the semantics, especially for tails, to the same level as heads.

We take one-hot encoding and sum aggregation as an example.
Here, $\mathcal{H}_l$ shows the 1-hop neighbor structures, and $\mathcal{H}_g$ gives the number of each class in the context.
Since $|\textit{MP}_{head}| \gg |\textit{MP}_{tail}|$, heads limit the feature spaces via a shorter context $\textit{GS}_{head}$, which nearly equals learning representation with only head data.
Meanwhile, we let $|\textit{GS}_{tail}| \approx N \gg |\textit{GS}_{head}|$, which accumulates node features in a subnetwork directed by serialization order rather than edges, in other words, message passing.
Thus, it covers a range of nodes with similar characteristics defined by $\sigma$ without losing its individuality when classifying tail data.
\paragraph{Framework with Graffin plugged.}
Figure~\ref{fig:framework}(D) shows how Graffin works when plugged.
There are two main streams of structural information flow.
Given a graph attribute matrix $\mathcal{X} \in \mathbb{R}^{N \times D}$, we keep local structures $\mathcal{H}_l$ of 1-hop neighbors through MP, using \textit{ReLU} as the activation:
\begin{equation}
    \mathcal{H}_l = \textit{ReLU}(\textit{MP}(\mathcal{X})),
    \label{eq:MPNNs}
\end{equation}
where \textit{MP} can be any message-passing framework aggregating neighbors in the same format as Equation~\ref{eq:MP}.

Meanwhile, we leverage GS to rearrange the node semantics as Equation~\ref{eq:GS} and \textit{Graffin} to control the information flows from heads to tails according to Equations~\ref{eq:Graffin1}-~\ref{eq:Graffin3}.
After that, we have global structures $\mathcal{H}_g$ of sequence standing for tail data:
\begin{equation}
    \mathcal{H}_g = \textit{Graffin}(\textit{GS}(\mathcal{X})).
\end{equation}

Next, we fuse two parts using Hadamard Product to form the final node representation $\mathcal{H}_f$ under the combined effect of neighborhood and sequence information:
\begin{equation}
    \mathcal{H}_f = \mathcal{H}_l \odot \mathcal{H}_g.
    \label{eq:fin}
\end{equation}
According to Equation~\ref{eq:Graffin3}, it is a three-part mixup: initial node representation, local structure via MP, and global structure via GS.
All of them are original-graph-oriented with easy access.

Finally, we feed $\mathcal{H}_f$ to a single linear classifier to compute the label predictions $\bar{\mathcal{Y}} \in \mathbb{R}^{N \times K}$ with an example negative log likelihood loss:
\begin{equation}
    loss = -\frac{1}{N}\sum_{i=1}^N(y_i\log\bar{y}_i + (1 - y_i)\log(1 - \bar{y}_i)).
    \label{eq:loss}
\end{equation}
In summary, Algorithm~\ref{alg:Graffin} shows the pseudocode of Graffin.
\begin{algorithm}[t]
    \caption{\textbf{Graffin: The Proposed Framework}}
    \label{alg:Graffin}
    \textbf{Input}: the input graph $\mathcal{G} = \{\mathcal{V}, \mathcal{E}, \mathcal{X}, \Phi, \mathcal{Y}\}$; \\
    \textbf{Output}: the final node representation $\mathcal{H}_f$;
    
    \begin{algorithmic}[1]
        \STATE // GS Phase
        \STATE Compute $\mathcal{V}'$ and $\mathcal{X}'$ according to Equation~\ref{eq:GS};
        \WHILE{not converged or not exceeded \textit{epochs}}
            \STATE Normalize $\mathcal{X}'$;
            \STATE Compute $\mathcal{T}_1$ and $\mathcal{T}_2$ according to Equations~\ref{eq:Graffin1} and ~\ref{eq:Graffin2};
            \STATE Compute $\mathcal{H}_g$ according to Equation~\ref{eq:Graffin3};
            \STATE // MP Phase
            \STATE Compute $\mathcal{H}_l$ according to Equation~\ref{eq:MPNNs};
            \STATE // Fusion Phase
            \STATE Rearrange $\mathcal{H}_g$ from $\mathcal{V}'$ to $\mathcal{V}$;
            \STATE Compute $\mathcal{H}_f$ according to Equation~\ref{eq:fin};
            \STATE Update the model according to Equation~\ref{eq:loss};
        \ENDWHILE
        \RETURN $\mathcal{H}_f$
    \end{algorithmic}
\end{algorithm}

\section{Experiments}
\paragraph{Datasets.}
We use four widely used datasets from real-world scenarios, including Amazon computers, Amazon photo, Cora, and DBLP.
Table~\ref{tab:dataset} shows the statistics of them.
Among datasets, DBLP constructs a heterogeneous graph.
We simplify it where nodes are from two types: \textit{Author} and \textit{Paper}, and edges from \textit{Author-Paper}.
Then, we apply the classification task on nodes of type \textit{Author}.
\paragraph{Baselines.}
We consider six pluggable baseline models from two categories:
\textbf{(1) Vanilla GRL models}: GCN~\cite{ref:GCN}, GAT~\cite{ref:GAT}, and GraphSAGE (SAGE)~\cite{ref:SAGE}.
\textbf{(2) SOTA models for imbalanced learning}: GraphSMOTE (SMO)~\cite{ref:GraphSMOTE}, GraphENS (ENS)~\cite{ref:ENS}, and ImbGNN (Imb)~\cite{ref:Imb}.
We further introduce SOTA models in Appendix B.
\paragraph{Evaluation Metrics.}
Following previous work~\cite{ref:GraphSMOTE} on evaluating imbalanced node classification, we set three measurements: overall accuracy (ALL), average AUC-ROC (A.R.), and macro F1-score (F1).
Dominated by head data, ALL may give a limited evaluation of tail data.
A.R. gives a robust view of classification performance, not relying on a specific threshold, while F1 considers all classes by calculating the harmonic mean of Precision and Recall.
Furthermore, we also report tail data accuracy (LOW), computing all tail examples at once.
All metrics can thoroughly reflect the performance of models on tail data.
\paragraph{Implementation Details.}
We run all experiments on a 64-bit machine with NVIDIA GeForce RTX 3050 OEM GPU, 8GB.
As for all vanilla GRL models, we apply a single layer and linear as the core and classifier.
We use Adam~\cite{ref:Adam} as the optimizer, initializing the learning rate to 0.01, with a weight decay 5e-4.
As for other imbalanced learning SOTA models, we reproduce their open-source codes and keep the settings with them.
SMO uses SAGE as the core layer and adds mean absolute error with square error as the final loss function.
ENS uses GCN instead, training with cross-entropy.
Though Imb is for imbalanced graph-level tasks, it treats each graph as a supernode to apply high-level node classification.
Thus, it is easy to transfer from graph-level tasks to node-level.
We utilize a stacked 5-layer GIN~\cite{ref:GIN} to encode, a 2-layer MLP to classify, and the same target loss as Graffin.
Codes are available in Appendix C.
To make a fair comparison, we repeat all experiments five times to report the average with the same training epoch of 200.
\begin{table}[!t]
    \centering
    \begin{tabular}{lrrrr}
        \hline
         \textbf{Datasets} & \textbf{\# N} & \textbf{\# D} & \textbf{\# Cl} & \textbf{$R_{imb}$} \\
         \hline
         Amazon computers & 13,752 & 767 & 10 & 17.73 \\
         Amazon photo & 7,650 & 745 & 8 & 5.86 \\
         Cora & 2,708 & 1,433 & 7 & 4.54 \\
         DBLP & 18,385 & 334 & 4 & 1.61 \\
         \hline
    \end{tabular}
    \caption{The statistics of datasets. \# N, \# D, and \# Cl show the number of nodes, features, and node classes. $R_{imb}$ indicates the node class imbalanced rate of each dataset.}
    \label{tab:dataset}
\end{table}
\subsection{Main Results on Node Classification}
\begin{table*}[!t]
    \centering
    % \resizebox{\linewidth}{!}{
    \setlength{\tabcolsep}{1mm}
    \begin{tabular}{cc|cc|cc|cc|cc|cc|cc}
        \hline
         \textbf{Dataset} & \textbf{Meas}. & \textbf{GCN} & + Gf & \textbf{GAT} & + Gf & \textbf{SAGE} & + Gf & \textbf{SMO} & + Gf & \textbf{ENS} & + Gf & \textbf{Imb} & + Gf \\
         \hline
         % COMP
         \multirow{4}{*}{computers} & ALL & $89.9^{1.0}$ & $87.7^{0.9}$ & $89.5^{3.0}$ & $\textbf{89.5}^{1.2}$ & $88.8^{1.5}$ & $87.2^{0.7}$ & $52.9^{2.8}$ & $\textbf{63.4}^{3.2}$ & $83.7^{0.8}$ & $\textbf{85.6}^{0.2}$ & $43.2^{1.5}$ & $\textbf{86.0}^{0.5}$ \\
         & LOW & $96.2^{0.7}$ & $\textbf{98.0}^{0.7}$ & $96.8^{1.8}$ & $\textbf{98.7}^{0.8}$ & $91.8^{5.2}$ & $\textbf{97.3}^{1.7}$ & $4.7^{2.2}$ & $\textbf{40.4}^{11.1}$ & $95.0^{0.6}$ & $\textbf{96.0}^{0.4}$ & $0.0^{0.0}$ & $\textbf{98.4}^{0.4}$ \\
         & A.R. & $99.5^{0.1}$ & $\textbf{99.9}^{0.0}$ & $99.7^{0.3}$ & $\textbf{99.9}^{0.0}$ & $99.5^{0.2}$ & $\textbf{99.9}^{0.0}$ & $89.5^{0.7}$ & $\textbf{94.6}^{0.9}$ & $99.4^{0.0}$ & $\textbf{99.7}^{0.0}$ & $74.3^{2.4}$ & $\textbf{99.9}^{0.0}$ \\
         & F1 & $88.0^{3.3}$ & $\textbf{98.0}^{0.1}$ & $94.5^{3.2}$ & $\textbf{98.4}^{0.3}$ & $88.0^{5.0}$ & $\textbf{98.0}^{0.3}$ & $48.7^{3.0}$ & $\textbf{62.1}^{4.0}$ & $82.4^{0.8}$ & $\textbf{83.2}^{0.3}$ & $13.4^{2.6}$ & $\textbf{97.6}^{0.2}$ \\
         \hline
         % PHOTO
         \multirow{4}{*}{photo} & ALL & $93.5^{1.2}$ & $92.6^{0.5}$ & $93.0^{1.2}$ & $\textbf{94.4}^{0.4}$ & $92.7^{3.8}$ & $\textbf{92.8}^{1.0}$ & $69.8^{2.0}$ & $\textbf{80.4}^{3.4}$ & $92.4^{0.2}$ & $92.3^{0.2}$ & $32.3^{8.6}$ & $\textbf{91.7}^{0.9}$ \\
         & LOW & $71.7^{10.3}$ & $\textbf{95.0}^{1.0}$ & $88.9^{5.7}$ & $\textbf{95.7}^{2.2}$ & $79.0^{15.5}$ & $\textbf{92.1}^{1.2}$ & $6.8^{3.8}$ & $\textbf{53.2}^{9.1}$ & $86.2^{0.5}$ & $\textbf{93.6}^{0.6}$ & $0.0^{0.0}$ & $\textbf{94.0}^{1.5}$ \\
         & A.R. & $99.6^{0.2}$ & $\textbf{99.9}^{0.0}$ & $99.6^{0.2}$ & $\textbf{99.9}^{0.1}$ & $99.7^{0.2}$ & $\textbf{99.9}^{0.0}$ & $94.3^{0.2}$ & $\textbf{95.6}^{0.9}$ & $98.9^{0.0}$ & $\textbf{99.8}^{0.0}$ & $70.2^{8.9}$ & $\textbf{99.9}^{0.0}$ \\
         & F1 & $93.5^{1.4}$ & $\textbf{98.2}^{0.2}$ & $96.1^{2.1}$ & $\textbf{98.7}^{0.1}$ & $92.8^{1.0}$ & $\textbf{98.0}^{0.2}$ & $66.4^{1.8}$ & $\textbf{80.2}^{3.6}$ & $90.5^{0.3}$ & $\textbf{92.0}^{0.2}$ & $16.9^{6.6}$ & $\textbf{98.0}^{0.2}$ \\
         \hline
         % CORA
         \multirow{4}{*}{Cora} & ALL & $84.2^{1.1}$ & $83.7^{0.9}$ & $79.9^{2.1}$ & $\textbf{83.0}^{1.1}$ & $79.6^{0.8}$ & $\textbf{80.7}^{0.7}$ & $45.8^{3.9}$ & $\textbf{53.3}^{2.6}$ & $84.7^{0.2}$ & $83.6^{0.4}$ & $32.3^{8.6}$ & $\textbf{91.7}^{0.9}$ \\
         & LOW & $89.2^{2.1}$ & $\textbf{89.3}^{2.8}$ & $83.1^{2.5}$ & $\textbf{86.0}^{3.2}$ & $83.5^{2.5}$ & $\textbf{84.2}^{2.3}$ & $21.8^{4.8}$ & $\textbf{49.8}^{9.6}$ & $90.7^{0.7}$ & $\textbf{91.1}^{0.2}$ & $0.0^{0.0}$ & $\textbf{94.0}^{1.5}$ \\
         & A.R. & $98.7^{0.1}$ & $\textbf{98.7}^{0.1}$ & $97.7^{0.6}$ & $\textbf{98.4}^{0.2}$ & $98.5^{0.1}$ & $\textbf{98.6}^{0.1}$ & $81.1^{2.5}$ & $\textbf{83.3}^{1.4}$ & $98.9^{0.0}$ & $\textbf{99.1}^{0.0}$ & $70.2^{8.9}$ & $\textbf{99.9}^{0.0}$ \\
         & F1 & $90.3^{0.6}$ & $\textbf{90.4}^{0.5}$ & $88.1^{0.9}$ & $\textbf{89.6}^{0.2}$ & $88.1^{0.6}$ & $\textbf{88.7}^{0.4}$ & $44.3^{4.8}$ & $\textbf{51.9}^{3.6}$ & $83.8^{0.2}$ & $83.0^{0.3}$ & $16.9^{6.6}$ & $\textbf{98.0}^{0.2}$ \\
         \hline
         % DBLP
         \multirow{4}{*}{DBLP} & ALL & $47.7^{1.1}$ & $\textbf{78.3}^{0.3}$ & $58.0^{2.0}$ & $\textbf{77.9}^{0.2}$ & $67.0^{1.2}$ & $\textbf{78.7}^{0.3}$ & $30.0^{0.6}$ & $\textbf{30.3}^{0.9}$ & $46.1^{0.3}$ & $\textbf{77.9}^{0.9}$ & $27.6^{3.0}$ & $\textbf{77.4}^{0.3}$ \\
         & LOW & $37.2^{1.5}$ & $\textbf{72.1}^{1.5}$ & $49.4^{2.8}$ & $\textbf{71.0}^{2.0}$ & $61.2^{8.4}$ & $\textbf{70.0}^{1.9}$ & $28.1^{8.1}$ & $\textbf{38.8}^{7.6}$ & $28.7^{2.0}$ & $\textbf{65.2}^{1.6}$ & $22.8^{5.0}$ & $\textbf{70.0}^{2.9}$ \\
         & A.R. & $75.4^{0.5}$ & $\textbf{93.1}^{0.2}$ & $81.5^{1.9}$ & $\textbf{93.8}^{0.1}$ & $88.9^{0.6}$ & $\textbf{92.7}^{0.3}$ & $54.5^{2.2}$ & $\textbf{58.4}^{4.2}$ & $70.5^{0.2}$ & $\textbf{93.1}^{0.4}$ & $54.9^{2.7}$ & $\textbf{94.0}^{0.1}$ \\
         & F1 & $53.3^{4.1}$ & $\textbf{79.5}^{0.2}$ & $61.0^{1.9}$ & $\textbf{79.0}^{0.2}$ & $69.4^{1.0}$ & $\textbf{79.7}^{0.2}$ & $25.0^{1.5}$ & $\textbf{27.3}^{1.6}$ & $43.7^{0.4}$ & $\textbf{76.8}^{0.8}$ & $28.8^{4.3}$ & $\textbf{78.6}^{0.3}$ \\
         \hline
    \end{tabular}
    % }
    \caption{Performance comparison on node classification with baselines. We report mean scores ($mean$) and standard deviations ($dev$) in the format $mean^{dev}$. +Gf means the base model plugged with Graffin.}
    \label{tab:main}
\end{table*}
\begin{figure*}
    \centering
    \includegraphics[width=\linewidth]{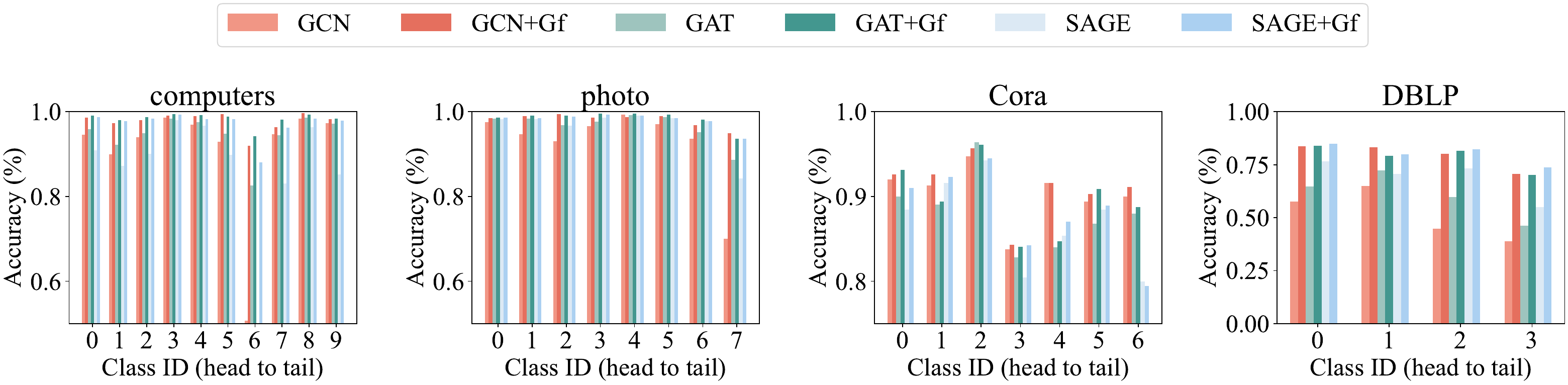}
    \caption{All classes accuracy w/o Graffin plugged. The class indexes are sorted from head to tail according to the number of nodes in each class.}
    \label{fig:class}
\end{figure*}
Table~\ref{tab:main} shows the main results of the node classification task.
All baseline models achieve noteworthy tail data performance gains with Graffin plugged.
From a dataset perspective, all baseline models perform poorly on DBLP.
Since we simplify it with only two types of nodes remaining, models can't capture enough features to learn the graph. 
However, with Graffin plugged in, all model performances improve by a huge step.
Graph sequence can dig out the latent structural information, finding long-distance node interactions to enrich tail representations.

Graffin successfully increases the accuracy of tail data, LOW, which is the initial purpose of our model.
The best case occurs when GCN plugs Graffin, which has an average LOW increase of 15.03\% among all four datasets.
We focus carefully on the balanced performance between classes, resulting in higher A.R. and F1 scores.
It shows that other nodes can benefit from the graph sequence, not only heads and tails.
Thus, the overall accuracy among all classes, ACC, also gains a lot, with only a few baseline models slightly dropping on one or two datasets.
We will show more details about the model performance of all classes later.

Surprisingly, Imb shows a competitive node classification ability with Graffin plugged, though it is for graph-level tasks at the beginning.
Imb goes from barely recognizing tail data (0\% LOW ACC for most datasets) to a classification level of around 90\%, illustrating a significant impact for the tails.
To sum up, Graffin can improve the adaptation to tail data without significantly degrading the overall model performance.
\subsection{All Classes Analysis}
\begin{figure*}[t]
    \centering
    \includegraphics[width=\linewidth]{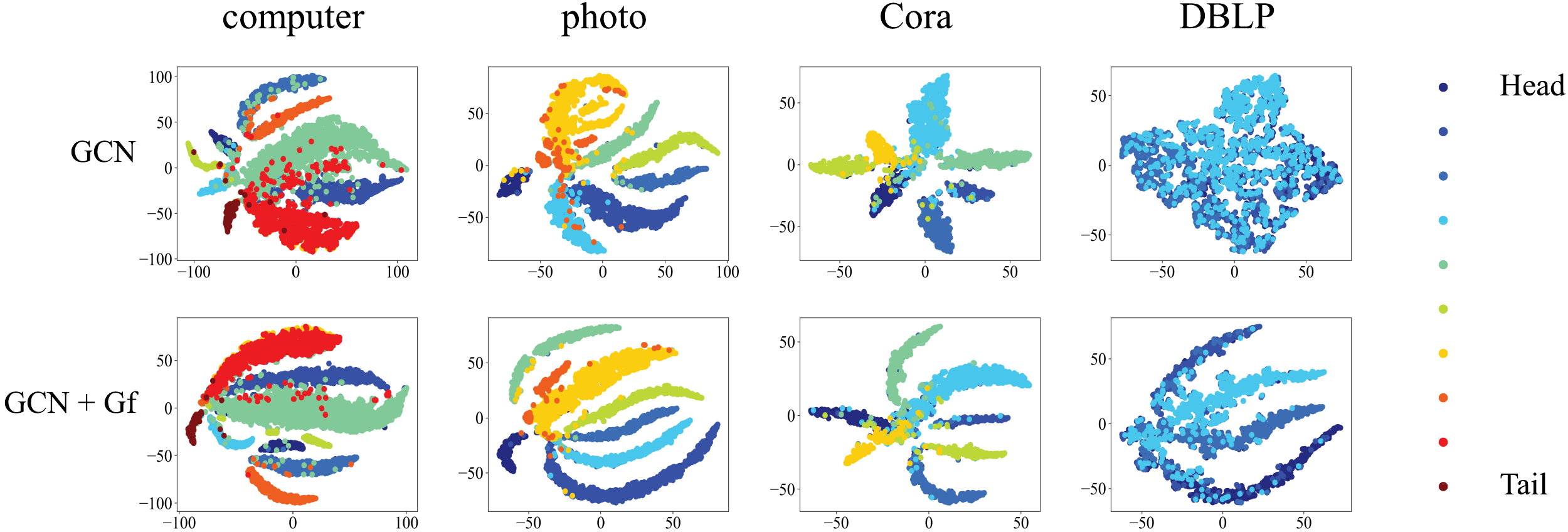}
    \caption{t-SNE visualization of GCN and GCN with Graffin (GCN+Gf) on all four datasets. Each color represents one node class from head data to tail data.}
    \label{fig:visual}
\end{figure*}
We further analyze the accuracy of all node classes in Figure~\ref{fig:class} to show how Graffin stands for tails in imbalanced node classification tasks.
In general, Graffin achieves its goal of improving the classification accuracy for tail data (the last bar in each subplot).
Cora has the least significant average growth of 0.44\%, while DBLP shows the best case of 24.88\%.
Head data also have a considerable increase.
% As in \textit{computer} and \textit{photo}, head performances increase by 5.01\% (5.00\% for tails) and 0.49\% (13.07\% for tails), respectively.
Although these nodes memorize shorter contexts in the sequence, they can still learn richer semantics due to their enormous base number.

As a matter of fact, except for heads and tails, we find other nodes benefit from the graph sequence, too.
An arbitrary node $v$ can learn more or less sequential information according to the serialization strategy, which complements the topological local neighbors.
Not all classes take advantage of the graph sequence.
The worst case is not precisely the tails.
For example, the worst performance occurring in \textit{computer} is 59.86\% for class 6 (93.19\% for tails), with a quantity ratio of 487:291.
% , in Cora is 82.36\% for class 0 (86.00\% for tails), with a ratio of 351:180.
That could explain why Graffin decays overall performances in some cases.
\subsection{Ablation on Graph Serialization Strategy}
\begin{table}[t]
    \centering
    % \resizebox{\linewidth}{!}{
    \setlength{\tabcolsep}{1mm}
    \begin{tabular}{cccccc}
        \hline
         \textbf{Dataset} & \textbf{Type} & \textbf{computer} & \textbf{photo} & \textbf{Cora} & \textbf{DBLP} \\
         \hline
         % GCN
         \multirow{3}{*}{GCN} & degree & $98.03$ & $98.23$ & $90.38$ & $79.48$ \\
         & eigen & -$0.05$ & -$0.03$ & -$0.42$ & -$0.03$ \\
         & id & -$0.08$ & -$0.03$ & -$0.77$ & -$0.74$ \\
         \hline
         % GAT
         \multirow{3}{*}{GAT} & degree & $98.37$ & $98.73$ & $89.62$ & $79.01$ \\
         & eigen & -$0.00$ & -$0.01$ & -$0.22$ & \underline{+$0.21$} \\
         & id & -$0.22$ & -$0.43$ & -$0.36$ & -$0.59$ \\
         \hline
         % SAGE
         \multirow{3}{*}{SAGE} & degree & $97.95$ & $97.97$ & $88.73$ & $79.72$ \\
         & eigen & -$0.24$ & \underline{+$0.02$} & -$0.68$ & -$0.81$ \\
         & id & -$0.25$ & -$0.74$ & -$1.42$ & -$0.95$ \\
         \hline
    \end{tabular}
    % }
    \caption{Ablation studies on different graph serialization strategies. We report relative F1 scores to consider precisions and recalls comprehensively.}
    \label{tab:ablation}
\end{table}
Except for node degrees, we also try two other serialization strategies: node eigenvector centrality and the default order of node ID.
The former computes $\lambda x^T = x^TA$ associated with the eigenvalue $\lambda$.
For node $v$, $\sigma = \sum_{u \rightarrow v} \sigma / \lambda$.
The latter is approximately equal to a basic random order.
Table~\ref{tab:ablation} illustrates the relative F1 scores of the ablation studies.
We set the node degree strategy as a baseline, where the negative number denotes the decline, and a positive number underlined denotes a better case.
Results show that the node degree strategy picked achieves higher and more stable scores.
The other two approaches are slightly weaker but still competitive concerning vanilla models.
The eigenvector is better than the node degree in GAT on DBLP (+0.21\%) and SAGE on \textit{photo} (+0.02\%).
The order of node ID, random order in other words, also provides considerable structural features within sequences.
The above conclusions reveal the effectiveness of the graph sequence in node representation, especially those breaking the graph topology, which can learn representations of nodes similar but not connected.
\subsection{Visualization}
To clearly illustrate the power of Graffin, we design visualization experiments via t-SNE~\cite{ref:t-SNE}.
We take GCN as an example due to the paper length limit.
Results of the other two vanilla models, GAT and SAGE, can be found in Appendix A.
As shown in Figure~\ref{fig:visual}, vanilla GCN can not distinguish tail data well, which mixes the representations of tails with other classes.
This phenomenon is more apparent in DBLP, where class representations couple tightly, resulting in underrepresented tail data.
GCN+Gf has more distinct boundaries among different classes.
With the help of sequential information, the representation space of each class shows the same form of fluid, which could match and evaluate with multiple linear.
Note there is a clustering center connected to each fluid where the hard samples fall.
GCN+Gf effectively reduces the scale of this area and the number of nodes in it.

\section{Conclusion}
In this paper, we aim to improve tail data performance in imbalanced node classification tasks without significantly corrupting the overall.
We propose a novel pluggable module, Graffin, enhancing topological semantics through graph sequences.
Graffin flows head features into tail data according to graph serialization order and learns the features along the sequence via RNNs to alleviate the imbalance of tail representation.
We fuse local and global structures to build a thorough view of neighbors and graph sequences.
We conduct the plug-play manner in six baseline models on four real-world datasets to demonstrate the effectiveness of Graffin.
The experimental results show that Graffin successfully adapts to tail data without significantly corrupting the overall model performance.

\bibliography{main}

\end{document}